\documentclass[letterpaper, 10 pt, conference]{ieeeconf}
\IEEEoverridecommandlockouts 
\usepackage{xcolor}
\usepackage{amsmath}
\usepackage{amssymb}  
\usepackage{cite}
\usepackage{multirow}
\usepackage{subfigure}
\usepackage{graphicx}
\usepackage{enumerate}
\usepackage[hidelinks]{hyperref}
\usepackage{booktabs}\usepackage{xcolor}

\DeclareMathOperator*{\argmin}{argmin}

\title{Learning-Augmented Model-Based Multi-Robot Planning for Time-Critical Search and Inspection Under Uncertainty}

\author{Abhish Khanal \and Joseph Prince Mathew \and Cameron Nowzari \and Gregory J. Stein \thanks{The authors are with
    the Department of Computer Science, George Mason University, VA,
    USA.} }

\begin{document}
\maketitle

\begin{abstract}
In disaster response or surveillance operations, quickly identifying areas needing urgent attention is critical, but deploying response teams to every location is inefficient or often impossible.
Effective performance in this domain requires coordinating a multi-robot inspection team to prioritize inspecting locations more likely to need immediate response, while also minimizing travel time.
This is particularly challenging because robots must directly observe the locations to determine which ones require additional attention. 
This work introduces a multi-robot planning framework for coordinated time-critical multi-robot search under uncertainty.
Our approach uses a graph neural network to estimate the likelihood of PoIs needing attention from noisy sensor data and then uses those predictions to guide a multi-robot model-based planner to determine the cost-effective plan.
Simulated experiments demonstrate that our planner improves performance at least by 16.3\%, 26.7\%, and 26.2\% for 1, 3, and 5 robots, respectively, compared to non-learned and learned baselines.
We also validate our approach on real-world platforms using quad-copters.

\end{abstract}

\section{Introduction} \label{sec:Intro}

In scenarios like disaster aftermath inspection or critical surveillance operations, quickly traveling to and inspecting affected areas is crucial for an efficient response. However, deploying a response team to every location can be both time-consuming and resource-intensive, especially since not all locations may be damaged or require immediate attention.
To address this, we consider a problem of deploying an agile multi-robot team to conduct a surveillance mission.
The goal is to have the robots search and inspect a predefined set of Points of Interest (PoI) in minimum expected cost.
The robots will identify which locations require urgent action, allowing the response team to be dispatched only to those areas needing immediate attention.

The challenge in these scenarios lies in coordinating multiple robots to quickly search for the locations that require immediate attention, since, without direct investigation, it’s unknown which locations will require further response.
The planner coordinating the robots must decide how to distribute their search efforts, balancing between prioritizing locations with the highest likelihood of being affected, choosing the nearest locations, or minimizing travel distance by covering all PoIs efficiently.
Effective coordination is essential to quickly search and report affected PoIs, as a delay in search and inspection could lead to loss of life or security breaches.

Classical approaches to multi-robot planning can be applied in a rapid inspection scenario by making simplifying assumptions, such as \emph{optimism under uncertainty}, which assumes that all locations might need immediate response. 
An \emph{optimistic} planner might prioritize the closest PoIs, leading to increased time to discover the locations that require immediate response.
Similarly, route optimization techniques like the Traveling Salesman Problem (TSP) \cite{hoffman2013traveling} and its multi-robot variant (mTSP) \cite{cheikhrouhou2014move, trigui2018analytical} are designed for operation in known environments and so focus on minimizing overall travel distance or time without regard to which locations might require more immediate inspection.
It is unclear how to directly apply these approaches to scenarios where it is uncertain which locations require immediate attention.

Instead, performing well requires inference of which locations require immediate attention, allowing the multi-robot team to prioritize those locations for inspection.
Noisy observations from static sensors monitoring the environment and historical data can provide valuable insights to infer which locations need priority.
Learning has the potential to leverage such observations, historical data, and the partial state of the world to infer which points of interest to prioritize the search \cite{wherewilleventoccur}. 
Reinforcement Learning (RL) based approaches, for example, can directly learn a multi-robot policy from noisy observations of the environment \cite{wei2021multi, dutta2021multi, gu2004multiagent}.
However, sparse or delayed rewards, characteristic of this problem setting, and the large multi-robot state space can lead to scalability challenges in practice that limit performance \cite{kober2013reinforcement}. 
As such, we need an approach that is capable of using learning to judge the goodness of searching different locations while also integrating the output of learning with a multi-robot planner.

\begin{figure}[]
    \centering
    \includegraphics[width=\linewidth]{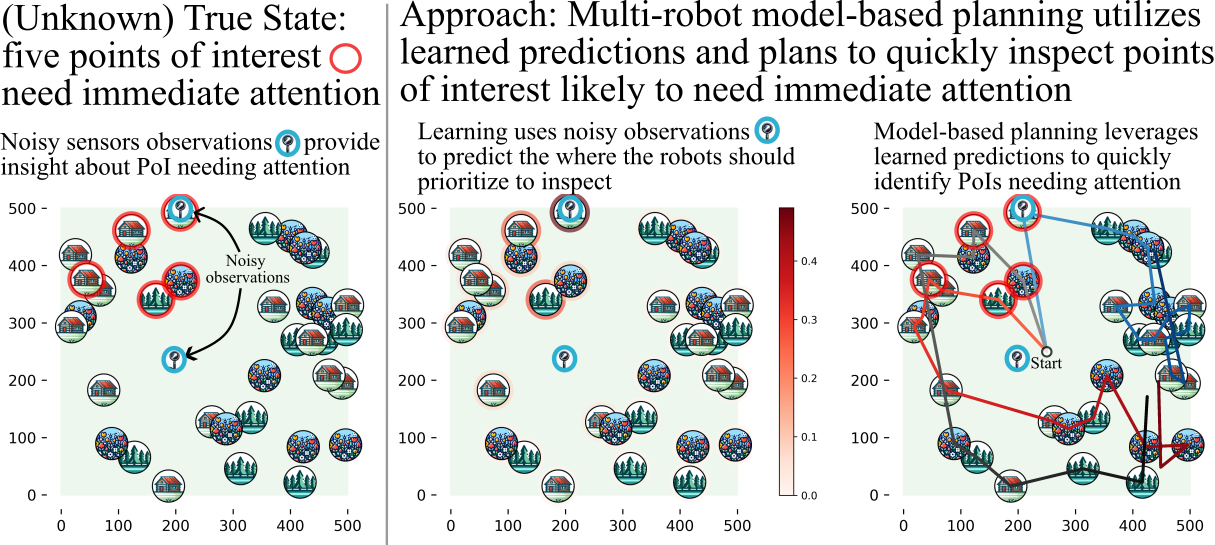}
    \caption{\textbf{An overview of our scenario and approach} Learning informs multi-robot model-based planning for effecting planning for time-critical search and inspection under uncertainty.}
    \vspace{-1.5em}
    \label{fig:intro}
\end{figure}

In this work, we introduce a multi-robot learning-informed model-based planning for this uncertain inspection scenario.
The core contribution of our approach is a model for multi-robot planning whose parameters---specifically, the transition probabilities representing the likelihood that certain locations require immediate attention---are estimated by learning.
We estimate these probabilities using a graph neural network that takes noisy sensor observations and partial state of the world as input.
Our multi-robot planning model computes a cost-effective policy for the team
which balances prioritizing locations that are prone to be affected, with checking nearby PoIs that can be quickly searched while en route to higher-priority locations.

We show the effectiveness of our learning-informed planner in procedural environments with 12 and 36 points of interest spread out across an open map for up to 5 robots. In simulated experiments, we show that our planner, guided by the predictions of our learned model, improves average performance relative to non-learned and learned baseline planners at least 16.3\%, 26.7\%, and 26.2\% for 1, 3, and 5 robots, respectively. We showcase the real-world applicability of our planner by deploying it on two quad-copters.

\section{Related Works}

Multi-robot planning has been extensively researched, with applications in disaster management \cite{ghassemi2022multi}, search and rescue \cite{maza2007multiple}, surveillance\cite{delle2009multi}, and agriculture\cite{ju2022review}.
In this work, we require coordinating multiple robots to efficiently search and inspect a set of locations to identify the locations that require immediate attention.

Classical approaches to multi-robot planning focus on task allocation \cite{faigl2015comparison, ghassemi2022multi} and route optimization in known, deterministic environments.
A common example is the multi-robot Traveling Salesman Problem (mTSP), which aims to minimize objectives such as travel distance or energy consumption by distributing tasks among the robots \cite{trigui2018analytical, cheikhrouhou2014move, wang2013novel}.
While effective in known environments, it is unclear how to search for unknown locations requiring immediate attention in minimum time without direct observation.


Other classical methods \cite{yamauchi1998frontier, bhattacharya2012topological, topological-mr}, which utilize abstractions that enable robots to visit multiple locations concurrently, can be used in the search and response scenario by making assumptions about the unseen space.
A common assumption is \emph{optimism under uncertainty} \cite{bhattacharya2015persistent}, which assumes all locations will require immediate attention.
As a result, the planner might prioritize nearby locations, potentially delaying the discovery of distant events that may require immediate attention, leading to inefficiencies.
A recent work in persistent UAV surveillance optimize visit sequences using Mixed Integer Linear Programming (MILP) to minimize the maximum weighted time between visits \cite{kalyanam2018scalable}.

Learning-based methods have the potential to bridge this gap by capturing the relationship between the state of the environment and sensor observations to infer which locations need urgent attention \cite{wherewilleventoccur}.
Reinforcement learning (RL) approaches \cite{wei2021multi, dutta2021multi, gu2004multiagent}, for instance, can learn multi-robot policies from noisy observations.
However, sparse and delayed rewards pose practical challenges, especially in large-scale multi-robot systems \cite{kober2013reinforcement}.

In this work, we present a multi-robot model that leverages learning to predict which locations require immediate attention and leverages this learned information to guide the search for the multi-robot team.



\section{Approach: Multi-robot Learning-Informed Model-Based Planning Under Uncertainty}
In this section, we introduce our multi-robot model-based planner for time-critical search and inspection scenarios. In Section~\ref{sec:Problem_formulation}, we formulate the cost model and use it with an MDP framework to compute the multi-robot team's joint action. Section~\ref{sec:learning} describes how learning is used to inform the planner. Finally, Section~\ref{sec:implementation} discusses the implementation details of the planner. 

\subsection{Problem Formulation} \label{sec:Problem_formulation}
We consider the problem of deploying a robot team for a surveying mission, where they must quickly search through a set of pre-defined locations $\mathcal{L}$ called Points of Interest (PoIs), and identify those that need immediate attention, minimizing the expected time.

Each point of interest is a discrete location denoted by $l \in \mathcal{L}$, which may require search and rapid response\footnote{In practice, these locations may be continuous; however, Voronoi cell decomposition can be used to represent the environment as discrete search points \cite{hu2020voronoi, voronoicontinuousuav, voronoi1}, partitioning the environment into distinct regions}, 
and is associated with a likelihood of need of secondary response $P(l)$ and 
an inspection time $T(l)$---the time required to inspect that location.

This section describes how cost is defined in Sec.~\ref{sec:cost-formulation} and the planning model for a multi-robot team that seeks to minimize the expected cost in Sec.~\ref{sec:planning}.


\subsubsection{Cost Formulation} \label{sec:cost-formulation}

The robot team needs to search through a set of predefined locations $\mathcal{L}$ to identify locations that need immediate attention.
There exists a cost per unit time $K$ for every time that 
accumulates whenever a location requiring immediate secondary attention goes un-inspected.
If $\Delta t$ is the amount of time that has passed while $N$ locations that need immediate attention are not discovered and reported, then the total cost can be written as $K \Delta t N$. 

The environment is only partially observed; therefore, the robot team does not know which of the points of interest need immediate attention.

We assume that for each point of interest location $l$, there exists an underlying probability $P(l)$ that reflects that the location needs immediate attention.
Hence, the total \emph{expected} cost incurred $\mathcal{C}$ during the passage of time $\Delta t$ is given by:  
\begin{equation}\label{eqn:cost-formulation}
    \mathcal{C} =  K \Delta t \sum_{l \in \mathcal{L}}P(l)
\end{equation}

\subsubsection{Planning for Multi-robot under Partial Observability} \label{sec:planning}
Though the uncertainty inherent in this problem means that in general it corresponds to a partially observable Markov decision process (POMDP), it is our insight that this problem can be represented as an MDP in which the costs associated with the team's actions instead correspond to an expected cost consistent with Eq.~\eqref{eqn:cost-formulation}.
In the general MDP formulation, the optimal state-action cost can be written as shown below:
\begin{equation} \label{eq:MDP}
Q(s_t, a_t) = R(s_t, a_t) + \min_{a_{t+1} \in \mathcal{A}(s_{t+1})} Q(s_{t+1}, a_{t+1})
\end{equation}
Where $s_t$ is the state, $a_t$ is the action, and $R(s_t, a_t)$ is the cost accumulated by taking action $a_t$ in state $s_t$.

Under our formulation, the state is defined as a tuple  $s_t = \{\mathcal{L}, \mathbf{q_t} \}$, where $\mathcal{L}$ is the set of un-inspected PoIs, and $\mathbf{q_t}$ is the list of robot locations.
For the state $s_t$ the set of joint actions $\mathcal{A}(s_t)$ involve assigning each robot a PoI to travel to and inspect, where $\mathcal{A}(s_t) =\bigotimes_{i\in I}(\mathcal{L})$ is constrained so that no two robots are assigned the same PoI whenever possible.
A single joint action $a_t \in \mathcal{A}$ can be represented as $a_t = [l_1, l_2 \cdots l_I]$, a list of target PoI locations, and consists of each robot moving to a specific point of interest and inspecting the location, where $l_i \in \mathcal{L}$ and $I$ is the total number of robots.



After the robot team is assigned a joint action, one robot finishes searching its assigned PoI before the others in the team and reports to the dispatch team if the PoI needs immediate attention.
We leverage insights from the learning-augmented multi-robot planning abstraction of Khanal et al.\cite{mrlsp}, which develops a similar framework for multi-robot indoor point-goal navigation, drawing inspiration from their approach.
Specifically, temporally extended actions in their work, as in ours, terminate when any robot has finished its assigned task. 
For our case, the expected time it takes for the robot $r$ to finish searching a PoI $l$ if it needs immediate attention is the sum of travel time $T_{\text{travel}}(q \rightarrow l)$ from the robot's current position ($q$) to the PoI ($l$), and the inspection time $T_R(l)$ of that PoI: i.e., $T_{\text{travel}}(q \rightarrow l) + T_R(l)$. Hence, for a joint-action, the first PoI that is inspected $L(s_t, a_t)$ and the time taken $T(s_t, a_t)$ can be written as shown in Eq.~\eqref{eq:time-and-poi}.
\begin{equation}\label{eq:time-and-poi}
\begin{split}
T(s_t, a_t) &= \min_{l \in a_t, q \in \mathbf{q_t}}\!\Big[T_{\text{travel}}(q \rightarrow l) + T_R(l)\Big] \\
L(s_t, a_t) &= \argmin_{l \in a_t, q \in \mathbf{q_t}}\!\Big[T_{\text{travel}}(q \rightarrow l) + T_R(l)\Big] 
\end{split}
\end{equation}

\begin{figure}[t] 
  \vspace{0.6em}
  \includegraphics[width=8.5cm]{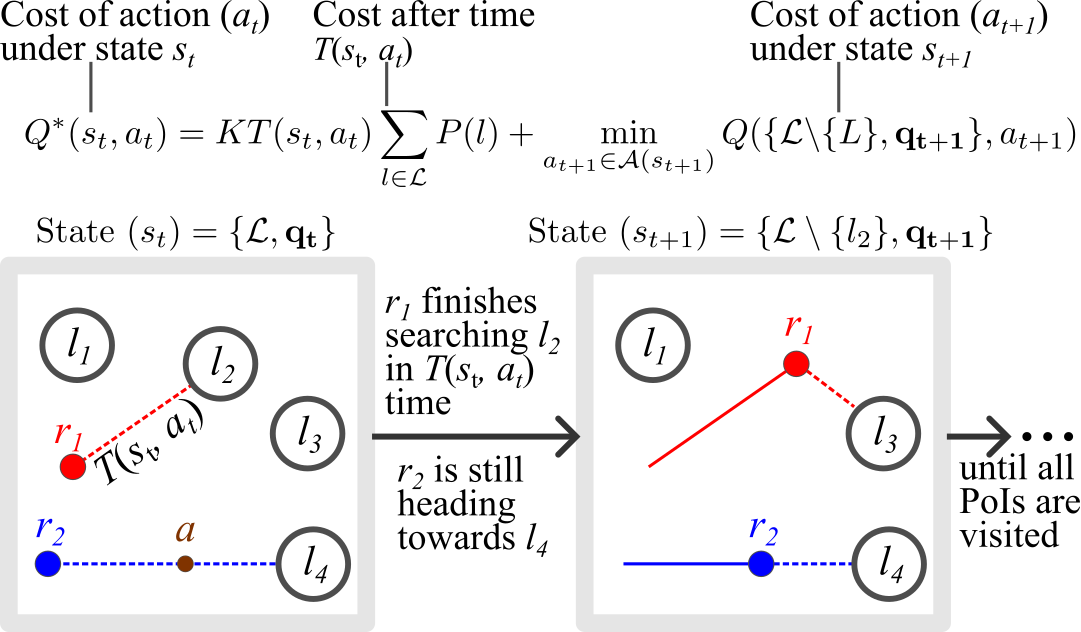}
  \caption{\textbf{Schematic of our Multi-robot Planning Framework}\quad{}A schematic showing how the cost of joint-action under a policy is calculated using our planning framework. For action $a_t$, the robot team concurrently explores multiple PoIs until one of the robots finishes inspecting a PoI, leading to a transition in the state.  \vspace{-1.5em}}\label{fig:mrlsp-schematic}
\end{figure}

After the first PoI $L$ is done inspecting, the state gets updated to $s_{t+1} = \{\mathcal{L} \backslash \{L\}, \mathbf{q_{t+1}}\}$ reflecting that $L$ is already inspected and removed from the set of points of interest $\mathcal{L}$, and the robots are in new locations $\mathbf{q_{t+1}}$.
The cost $R(s_t, a_t)$ in Eq.~\eqref{eq:MDP} accumulates over the duration $T(s_t, a_t)$ that passes during the state transition from $s_t$ to $s_{t+1}$ and is computed using Eq.~\eqref{eqn:cost-formulation}. The expected state-action cost in Eq.~\eqref{eq:MDP} can then be rewritten as:
\begin{equation}
\begin{split}\label{eq:planning-unknown}
    Q^*(s_t&=\{\mathcal{L}, \mathbf{q_t}\}, a_t) = K T(s_t, a_t) \sum_{l \in \mathcal{L}}P(l) \\
     & + \min_{a_{t+1} \in \mathcal{A}(s_{t+1})} Q(\{\mathcal{L} \backslash \{L\}, \mathbf{q_{t+1}}\}, a_{t+1}) 
\end{split}
\end{equation}
where $K$ is a constant denoting cost per unit time, $T(s_t, a_t)$ is computed using Eq.~\eqref{eq:time-and-poi} and $P(l)$, the likelihood that a PoI needs immediate attention, is estimated via learning.
\subsection{Estimating the Likelihood via a Graph Neural Network} \label{sec:learning}
Since the likelihood each location will require secondary response will not be known to the team, they must estimate this likelihood from noisy sensor data, a task for which learning has proven to be an effective tool.

We want our learning to be flexible across range of environments and scenarios with varied number of points of interest and observations.
A graph neural network (GNN) is well-suited for this task.
As such, we represent the environment as a graph where nodes correspond to points of interest and edges represent spatial proximity. 
Thus the graph neural network can consume this graphical representation allowing it to infer which locations are most likely to require attention based on historical patterns and noisy sensor observations.

In our disaster response scenario, the GNN estimates the likelihood that specific locations (such as buildings, forests, or fields) have been damaged based on noisy sensor data.
For example, Doppler radar data detecting pockets of high-wind gusts can provide an indication of areas that might have suffered damage, and historical information on past storm impacts can further inform the likelihood of damage at different types of locations.
By incorporating both current sensor readings and prior knowledge, the GNN predicts which locations are most likely to require immediate response to inform planning via Eq.~\eqref{eq:planning-unknown}.

In general, the type of noisy observation depends on the scenario: for instance, in a secured perimeter surveillance, noisy data could come from motion sensors or cameras tracking infiltrators’ last known locations, while in environmental monitoring, satellite imagery or sensor readings could predict high-probability locations for resources.

\subsection{Planner Details and Implementation} \label{sec:implementation}
Our planning framework has two components: the planner itself, which seeks to solve Eq.~\eqref{eq:planning-unknown}, and a learned model that is used to estimate the likelihood that any points of interest need immediate attention, needed to determine the lowest expected cost plan. 

The planner takes as input the list of robot poses and the list of points of interest, which includes (1) their location, (2) the (estimated) probability that the PoI needs immediate attention and (3) the (known) amount of time it would take to inspect the location $T_R(l)$.
The planner also takes the distances $T(q\xrightarrow{}l)$ from the robot's current pose  $q \in \mathbf{q_t}$ to each PoI $l \in \mathcal{L}$ and inter-distances between each PoI.
We implement the optimization procedure implicit in Eq.~\eqref{eq:planning-unknown} via a depth-first search and branch-and-bound. Our implementation is in C++ with Python bindings made possible via PyBind11 \cite{pybind11}, allowing both speed and easy connection to plotting and visualization tools.

Even using branch-and-bound to accelerate search, the speed of planning scales poorly with the number of robots and points of interest, since the number of feasible plans scales factorially with the increase in the number of robots and locations to visit.
To circumvent this computational challenge, we implement a \emph{receding horizon} planning strategy, in which we pick a subset of the points of interest $\mathcal{L_{\text{priority}}} \subseteq \mathcal{L}$ and instead find the lowest expected cost plan using only that subset via Eq.~\eqref{eq:planning-unknown}: $Q(\{\mathcal{L}_{\text{priority}},q\},a)$. Whenever one of the robots reaches a location, revealing it and reporting to the dispatch team if the locations require immediate attention, the whole team replans using the list of unrevealed locations.

To select the prioritized points of interest list $\mathcal{L}_{\text{priority}}$, we use a simple heuristic, selecting first the PoIs closest to each robots and then the most likely, up to a total of $N_{\text{PoI}} \equiv |\mathcal{L}_{\text{priority}}|$ locations. In all experiments, we use $N_{\text{PoI}} = 12$, selecting the first 6 PoIs most likely to need immediate response, and the remaining 6 PoIs nearest to each robot in order. This heuristic ensures that the robots search nearby locations when advantageous, while prioritizing locations likely to need immediate attention.

\section{Experimental Results}
\begin{figure}[t]
    \centering
    \includegraphics[width=8.5cm]{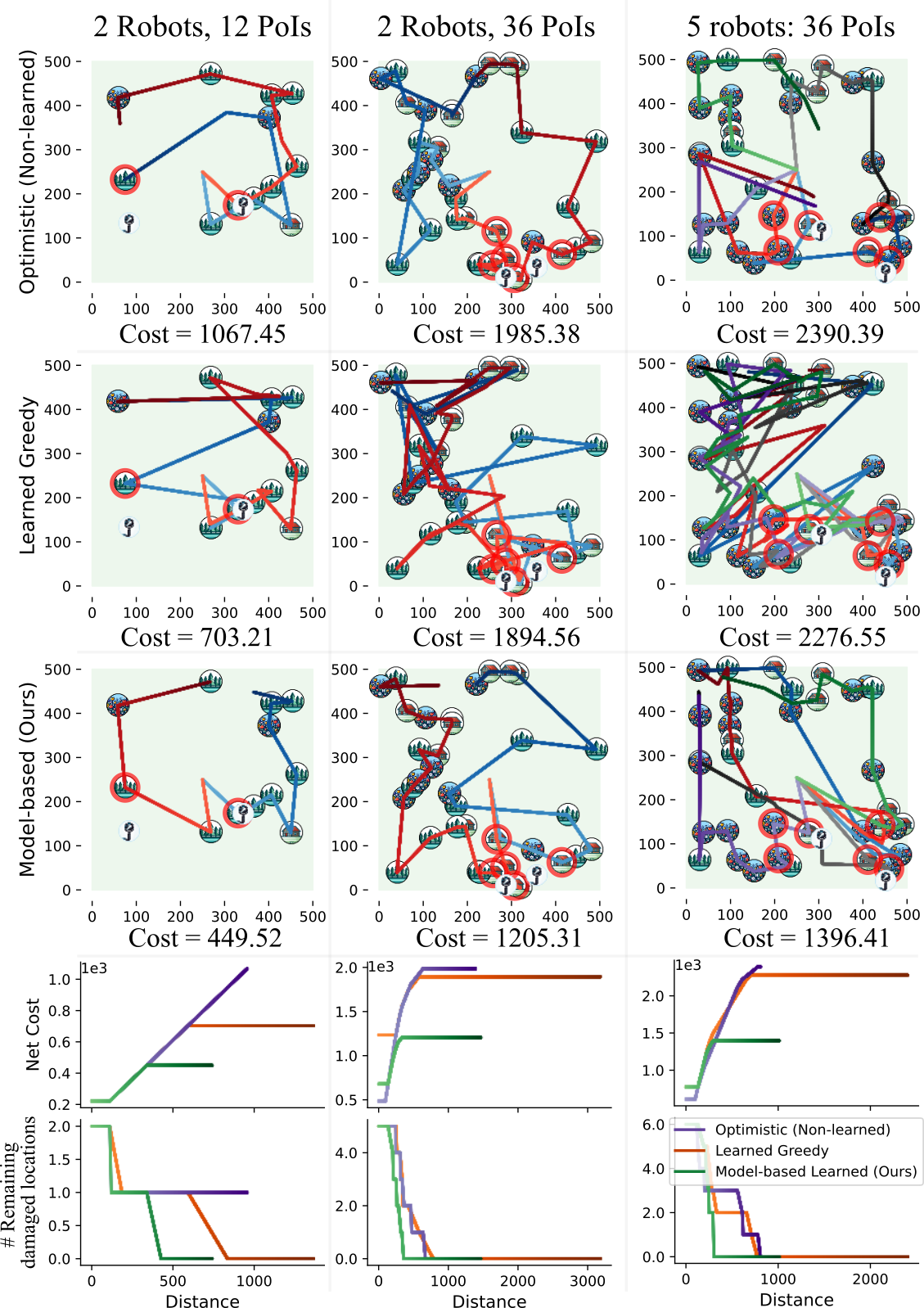}
    \caption{\textbf{Example trials}\quad We show the trajectory for optimistic, learned greedy, and our approach in a simulated damage survey scenario. The bottom plot shows distance vs net cost and remaining number of damaged locations for each planner.}
    \vspace{-1.5em}
    \label{fig:detailed_result}
\end{figure}
For the experiments, we consider the problem of deploying a UAV in a simulated damage-surveying mission where a set of Points of Interest (buildings, fields, and forests) within an area need to be inspected after a storm.
The area is monitored with sensors, such as Doppler radar, which detect pockets of high wind gusts. The UAVs are tasked with visiting a set of locations to inspect potential damage and report to a secondary response team if the location is damaged.

We present statistical results from a procedurally generated damage survey scenario and further demonstrate the real-world applicability of our planning framework through a physical experiment with a two-UAV team.

\begin{figure}[t]
    \centering
    \vspace{0.5em}
    \includegraphics[width=8.5cm]{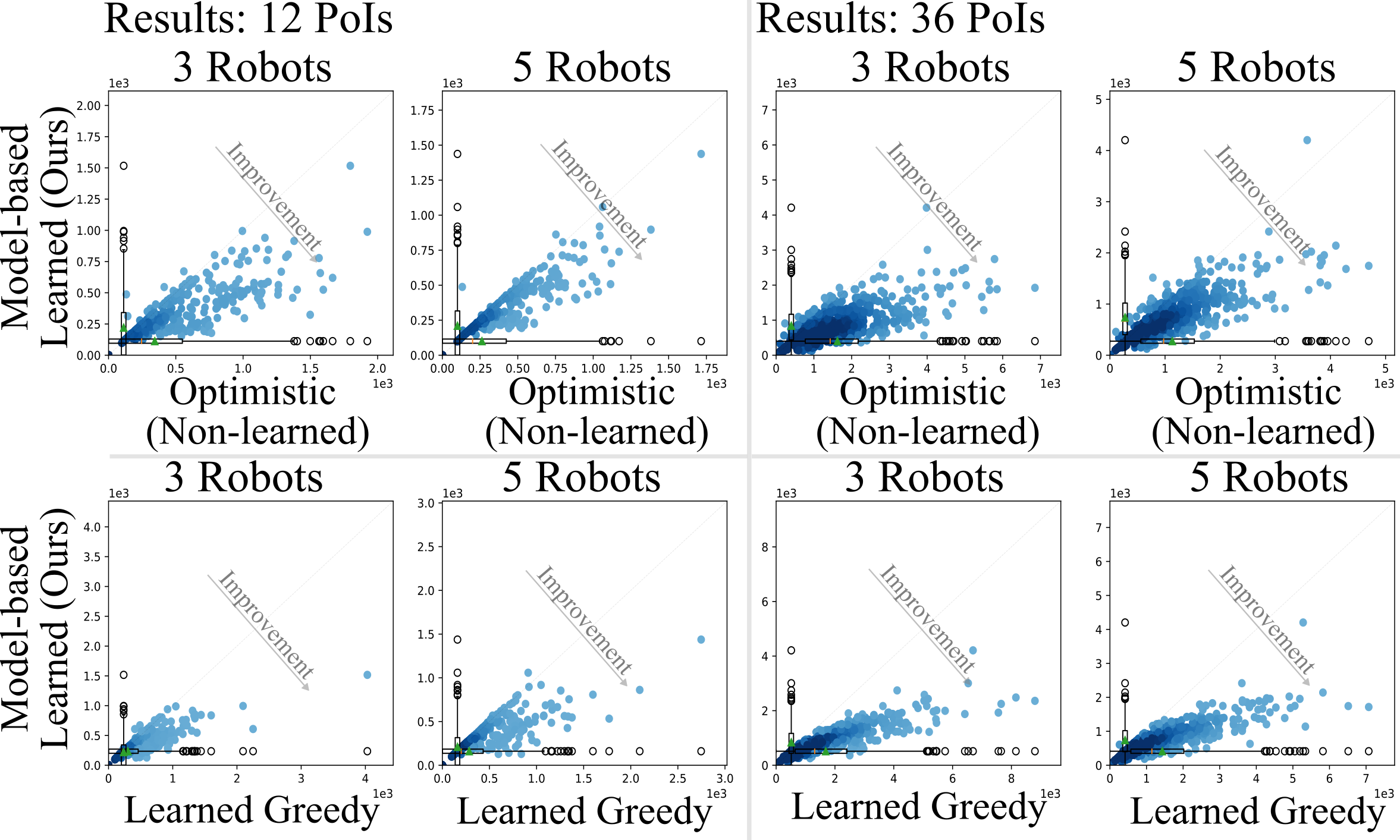}
    \caption{\textbf{Results: 12 and 36 PoIs} \quad  We compare the cost of our multi-robot model-based learned planner vs learned and non-learned baselines for 500 experiments in the scatterplot. The statistics for all the results show the benefit of our learning-augmented model-based planner.}
    \label{fig:scatter_result}
    \vspace{-1.5em}
\end{figure}

\subsection{Procedural Scenario Generation} \label{sec:procedural_scenario}
We conduct simulated experiments in procedurally generated environments corresponding to our damage-surveying scenario. The scenario consists of a set of points of interest $\mathcal{L}$ (buildings, forests, and fields), the locations of which are known to the robot team in advance.

To generate the environment procedurally, a number of points of interest and high-wind locations are randomly placed in the environment relative to the center of the map, from where the robot team starts the search. In this setting, the inspection time $R(l)$ is set to a constant value of 30.


The likelihood of damage for each PoI depends on how susceptible it is to damage, and its proximity to the high-wind location according to a Gaussian function:
\begin{equation} \label{eq:gaussian}
    P_{\text{damage}}(s, l) = P_{\text{susceptible}}(l)\text{exp}(-d^2(s, l)/2\sigma^2)
\end{equation}
where $s$ is the location of high-winds pocket, $l$ is the PoI, $d(s, l)$ is the distance between the location of high-wind pocket $s$ and the point of interest $l$, $\sigma = 60$ is the scale factor governing the influence range of the high-wind, and $P_{\text{susceptible}}$ is inherent vulnerability to damage of the PoI with values of 1.0 for forests, 0.8 for fields, 0.2 for buildings. In all experiments, two wind-pocket locations are generated and used. For each PoI, a Bernoulli random variable is sampled, using Eq.~\eqref{eq:gaussian}, to determine whether or not it is damaged.

\subsection{Data Structure and Training Details}
We use a procedurally generated disaster response scenario (as discussed in Sec.~\ref{sec:procedural_scenario}) to create data for training our model. Specifically, we generate 9,000 environments for each of 12, 24, and 36 points of interest, resulting in a total of 27,000 training samples. 
The data consists of sampled environments of our damage-surveying scenario.
The data label associated with each point of interest is whether or not it is damaged because of the storm.
We represent the environment as a graph with nodes $\mathcal{N}$ and edges $\mathcal{E}$, which are input into a graph neural network. Nodes $\mathcal{N}$ correspond to points of interest and wind gust pockets, with a feature vector as a one-hot encoding of the node's class: {field, forest, cabin, or wind gust pocket}. Edges are added when nodes are within 400 meters, with each edge having a float feature based on the distance: 1 for zero distance, 0 at the threshold, and linearly interpolated in between.

To use this data, we use a graph neural network architecture that handles both node and edge features, for which the TransformerConv layer is well-suited \cite{shi2020masked}.
Since edges in TransformerConv are directional, each undirected edge is replaced by two directed edges, one in each direction.
Additionally, we add self-edges to ensure that each node has a direct connection to itself.

Our network begins with four TransformerConv layers, each with an input node feature dimension of 4 and 16-dimensional hidden layers, followed by Leaky ReLU activation and 1D batch normalization.
The output is passed through a 16-to-1 linear layer, producing logits for the cross-entropy loss function and the predicted probability of a location needing attention via a sigmoid function. This network model is trained for 10 epochs with a learning rate of 5e-4. 

\subsection{Simulation Results}
We conduct batch simulation experiments across our procedurally generated scenario to evaluate the performance of our learned model-based planner with learned and non-learned planner baselines as described below:

\textbf{Learned Model-based Planner (ours)}: This planner uses learning to estimate the likelihood via our learned model and uses Eq.~\eqref{eq:planning-unknown} to effectively divide and conquer to inspect the points of interest.

\textbf{Non-learned Optimistic Planner} :
This approach guides the robot team to inspect Points of Interest (PoIs) based on proximity. Robots visit the nearest un-inspected PoI from their location, repeating this until all PoIs are inspected.

\textbf{Learned Greedy Planner}: This planner uses our learned model (Sec.~\ref{sec:learning}) to compute the likelihood that a point of interest needs immediate attention, and the robot team is assigned to the PoI with high likelihood needing a secondary response.

Each planner is evaluated in a previously unseen 500 unique simulated environment with 12, 24, and 36 PoIs for 1, 3, and 5 robots. Results are shown in various forms.

\begin{figure*}[t]
    \vspace{1em}
    \centering
    \footnotesize
    \begin{tabular}{c ccc ccc ccc}
    \toprule
                                         & \multicolumn{3}{c}{12 PoIs}                                            & \multicolumn{3}{c}{24 PoIs}                                            & \multicolumn{3}{c}{36 PoIs}                                            \\ 
    \cmidrule(lr){2-4}  \cmidrule{5-7}  \cmidrule(ll){8-10}                                
    Planner                              & \multicolumn{1}{c}{1 Robot} & \multicolumn{1}{c}{3 Robots} & 5 Robots & \multicolumn{1}{c}{1 Robot} & \multicolumn{1}{c}{3 Robots} & 5 Robots & \multicolumn{1}{c}{1 Robot} & \multicolumn{1}{c}{3 Robots} & 5 Robots \\ 
    \midrule
    Optimistic (non-learned)                              & \multicolumn{1}{c}{775.4}      & \multicolumn{1}{c}{343.7}       & 262.5       & \multicolumn{1}{c}{2126.2}      & \multicolumn{1}{c}{907.2}       & 635.8       & \multicolumn{1}{c}{3936.2}      & \multicolumn{1}{c}{1622.9}       & 1131.9       \\
    Learned greedy    & \multicolumn{1}{c}{435.1}      & \multicolumn{1}{c}{301.7}       &    284.9    & \multicolumn{1}{c}{1367.6}      & \multicolumn{1}{c}{821.4}       &         719.1    & \multicolumn{1}{c}{3092.3}      & \multicolumn{1}{c}{1685.6}       &        1431.2  \\
    
    Model-based (learned) & \multicolumn{1}{c}{\textbf{364.2}}      & \multicolumn{1}{c}{\textbf{221.1}}       &  \textbf{210.9}      & \multicolumn{1}{c}{\textbf{903.1}}      & \multicolumn{1}{c}{\textbf{493.5}}       &    \textbf{451.3}    & \multicolumn{1}{c}{\textbf{1719.0}}      & \multicolumn{1}{c}{\textbf{842.4}}       & \textbf{740.5}       \\
    \bottomrule
    \vspace{-1em}
    \\
    \end{tabular}
    \vspace{-0.5em}
\includegraphics[width=16cm]{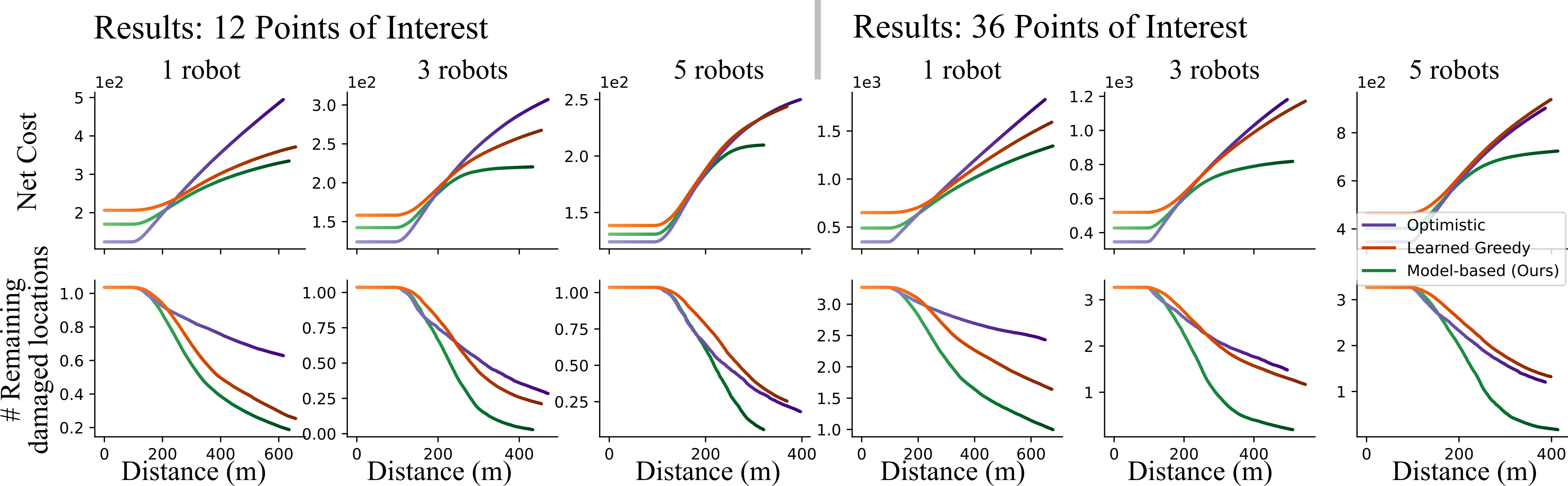}
\caption{\textbf{Results for 12, 24 and 36 PoIs}\quad The table shows the average cost accrued in 500 experiments for each planner. The plot net cost and number of remaining damaged locations vs. distance to visit all PoIs covered for our planner (green), learned-greedy baseline (orange), and optimistic baseline (purple). The results show our model-based planner finds all damaged locations quickly, accruing less cost compared to non-learned and learned baselines.}
\vspace{-1.3em}\label{fig:table}
\end{figure*}

\subsubsection{Results: Environments with 12 Points of Interest}
We first show performance in environments with 12 points of interest. The statistics, shown in Fig~\ref{fig:table} show that our learning-informed model-based planner outperforms the non-learned and learned baselines by 16.3\%, 26.7\%, and 26.2\% for 1, 3, and 5 robots.

More detailed results are shown in Fig.~\ref{fig:scatter_result}, in which we include scatter plots showing the relative performance of our learning-informed model-based approach against each baseline. Each point in the scatter plot corresponds to the cost incurred (using Eq.~\eqref{eqn:cost-formulation}) for our learning-informed model-based planner (y-axis) vs. the baseline planner (x-axis) in a single trial.
More points below the diagonal line show that our planner yields improvement in cost over both non-learned and learned baselines across all settings.

We also highlight a representative trial in Fig.~\ref{fig:detailed_result}(a), showing trajectories followed by each planner and cost over time for each. Our planner, informed by predictions from learning, identifies that dense wooded locations (forests) are more prone to damage by nearby storm centers and prioritizes travel in that region to search and inspect for damages. 

\subsubsection{Results: Environments with 24 and 36 PoIs}
To assess the scalability of our approach, we extended our experiments to more complex environments with 24 and 36 points of interest. Despite the increased challenge, our learning-informed planner still outperforms learned and non-learned baselines. The table in Fig.~\ref{fig:table} illustrates that even in these more intricate settings, our model outperforms the baselines by 34.0\%, 39.9\%, and 37.3\% for 1, 3, and 5 robots, respectively.

Fig.~\ref{fig:detailed_result} also presents scatter plots for 36 point-of-interest environments for 3 and 5 robots that show the advantage of our approach. The overall trend across all environment sizes demonstrates that our learning-informed planner consistently yields improvements over non-learned and learned baselines.

\subsection{Demonstration on a Real-World Platform}
\begin{figure}
    \centering
    \includegraphics[width=\linewidth]{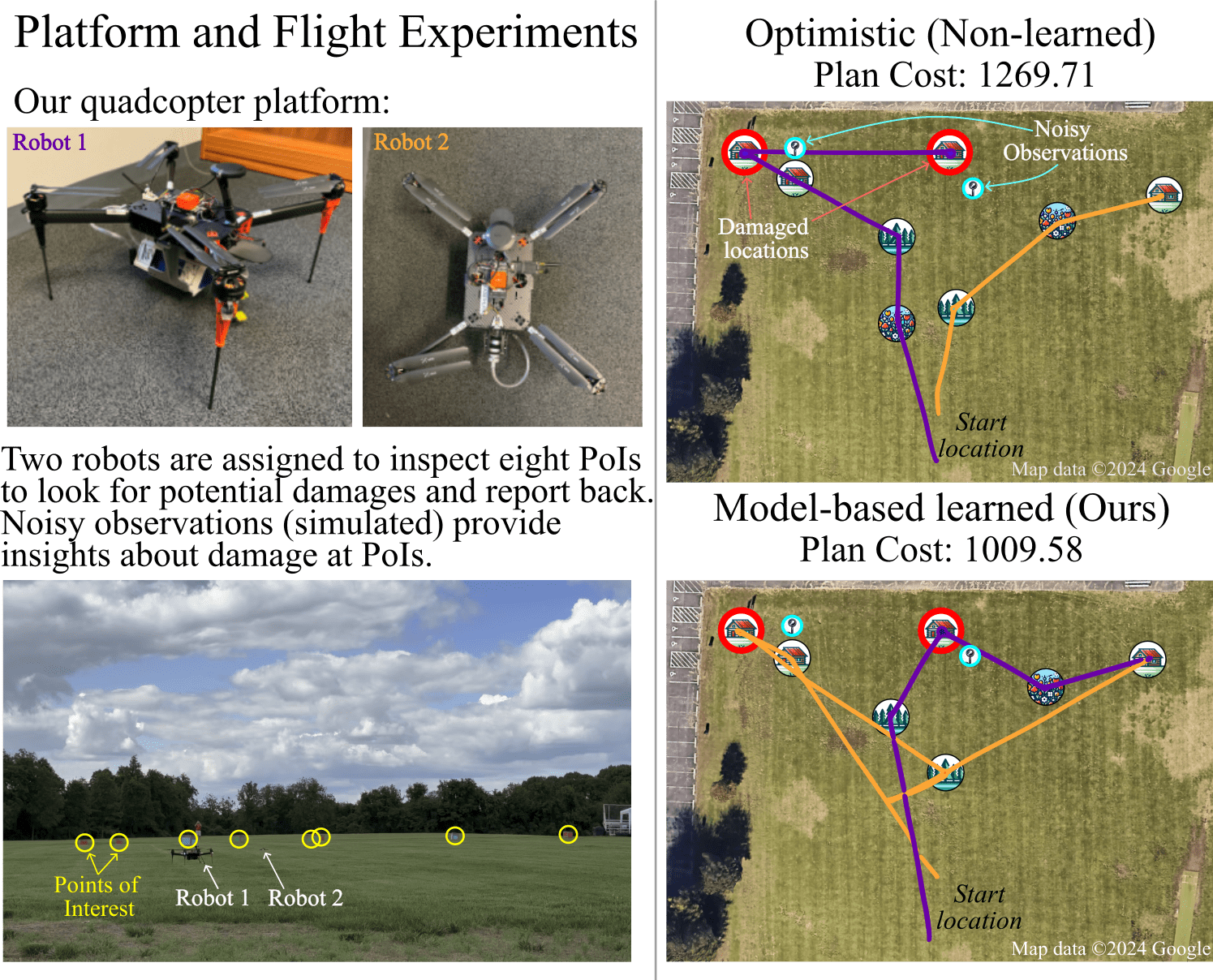}
    \caption{\textbf{Demonstration of feasibility in deployment of our planner in two quadcopters} \quad The figure shows our quadcopter platforms (top left). Bottom left shows the environment where the experiment is done, where PoIs are boxes, circled in yellow. These PoIs denote locations that need to be searched for survivors. (Top right): Trajectory from optimistic planner. (Bottom right): Trajectory from our learning-augmented multi-robot model-based planner. We show the feasibility of our approach through these physical experiments.}
    \label{fig:realworld}
    \vspace{-1.5em}
\end{figure}
To validate the effectiveness and feasibility of our approach in real-world scenarios, we deployed both our learning-informed planner and a greedy planner on a two quad-copter team. These unmanned aerial vehicles utilize the Ardupilot flight stack as their low-level motion controller, enabling precise navigation between waypoints that correspond to points of interest. Each quad-copter is equipped with a Global Positioning System (GPS), a 9 degree-of-freedom Inertial Measurement Unit (IMU), and a barometer, collectively providing accurate 3D state estimation. An onboard Raspberry Pi-4 facilitates communication with a centralized computer, receiving waypoint instructions and transmitting estimated 3D state information.

Using our procedural environment generation method, we created a test scenario comprising 10 points of interest for the quad-copters to navigate. The centralized computer, using real-time pose data from the quad-copters and the coordinates of all points of interest, computes plans and transmits waypoint instructions to the vehicles. This process was carried out for both our learning-informed planner and the greedy baseline approach.

Fig.~\ref{fig:realworld} illustrates the flight trajectories of the two quad-copters under both planning strategies. The results demonstrate the effectiveness of our learning-informed model. Using our learning-informed approach, the quadcopters prioritize areas near simulated noisy observations and swiftly reach points of interest that need a secondary response team. In contrast, the optimistic planner conducts its search in a less informed manner, resulting in a 20.43\% increase in accumulated cost.

These real-world trials mirror the outcomes observed in our fully-simulated experiments, providing further evidence of our approach's viability for deployment on actual platforms. The successful implementation of our learning-informed planner on physical quad-copters underscores its potential for enhancing efficiency and responsiveness in various real-world applications, from disaster response, search and rescue or surveillance tasks.

\section{Conclusion}
In this work, we present a learning-augmented multi-robot planning framework for efficient search under uncertainty in scenarios such as disaster response and surveillance.
By using a graph neural network to estimate the likelihood of Points of Interest (PoIs) needing attention based on noisy sensor data, we integrated these predictions into a model-based planner that balances prioritization of high-risk locations with minimizing travel time. From statistical results, we present improvement in performance over learned and non-learned baselines, showing the efficacy of our learning-informed multi-robot model-based planner for up to 5 robots. 
However, as the number of robots and PoIs increases, the joint action space grows factorially, making planning computationally expensive. Nonetheless, the insights from our approach could be valuable in related domains. For instance, target tracking methods \cite{nowzari2012self} could benefit from learning-based likelihood predictions, while UAV-based persistent surveillance \cite{kalyanam2018scalable} could leverage learning to optimize node visit frequency. 


\section*{Acknowledgement}
This material is based upon work supported by the National Science Foundation under Grant No. 2232733 and by the Army Research Lab under award W911NF252001. The authors acknowledge support from these awards.

\bibliography{references}
\bibliographystyle{IEEEtran}

\end{document}